\documentclass[10pt,twocolumn,letterpaper]{article}

\usepackage{cvpr}
\usepackage{times}
\usepackage{epsfig}
\usepackage{graphicx}
\usepackage{amsmath}
\usepackage{amssymb}
\usepackage{caption}
\usepackage{authblk}
\usepackage[toc,page]{appendix}

\DeclareMathOperator*{\argmin}{arg\,min}

\cvprfinalcopy

\title{Compositional Temporal Visual Grounding of\\ Natural Language Event Descriptions}

\author[1]{Jonathan C. Stroud\thanks{Corresponding author. \texttt{stroud@umich.edu}}}
\author[2]{Ryan McCaffrey}
\author[1]{Rada Mihalcea}
\author[2]{Jia Deng}
\author[2]{Olga Russakovsky}

\affil[1]{University of Michigan}
\affil[2]{Princeton University}

\ifcvprfinal\fi
\begin{document}
\maketitle
\begin{abstract}
Temporal grounding entails establishing a correspondence between natural language event descriptions and their visual depictions. \emph{Compositional} modeling becomes central: we first ground atomic descriptions (``girl eating an apple,'' ``batter hitting the ball'') to short video segments, and then establish the \emph{temporal} relationships between the segments. This compositional structure enables models to recognize a wider variety of events not seen during training through recognizing their atomic sub-events. Explicit temporal modeling accounts for a wide variety of temporal relationships that can be expressed in language: \eg, in the description ``girl stands up from the table after eating an apple'' the visual ordering of the events is reversed, with first ``eating an apple'' followed by ``standing up from the table.'' We leverage these observations to develop a unified deep architecture, CTG-Net\footnote{Project page: \texttt{jonathancstroud.com/ctg}}, to perform temporal grounding of natural language event descriptions to videos. We demonstrate that our system outperforms prior state-of-the-art methods on the DiDeMo, Tempo-TL, and Tempo-HL temporal grounding datasets.
\end{abstract}

\section{Introduction}

We consider the task of temporally grounding natural language event descriptions in videos. In this task, we are given a video and a natural language query, and we must locate the point in time that corresponds to the query. For example, we may have a video of a baseball game, along with the query ``the batter hits the ball, and a player tags him out at first base.'' The system must find the point in time that best corresponds with the event described in the query. A system that can accomplish this task would have many applications, such as video retrieval and human-robot interaction.

\begin{figure}
\centering
\includegraphics[width=0.475\textwidth]{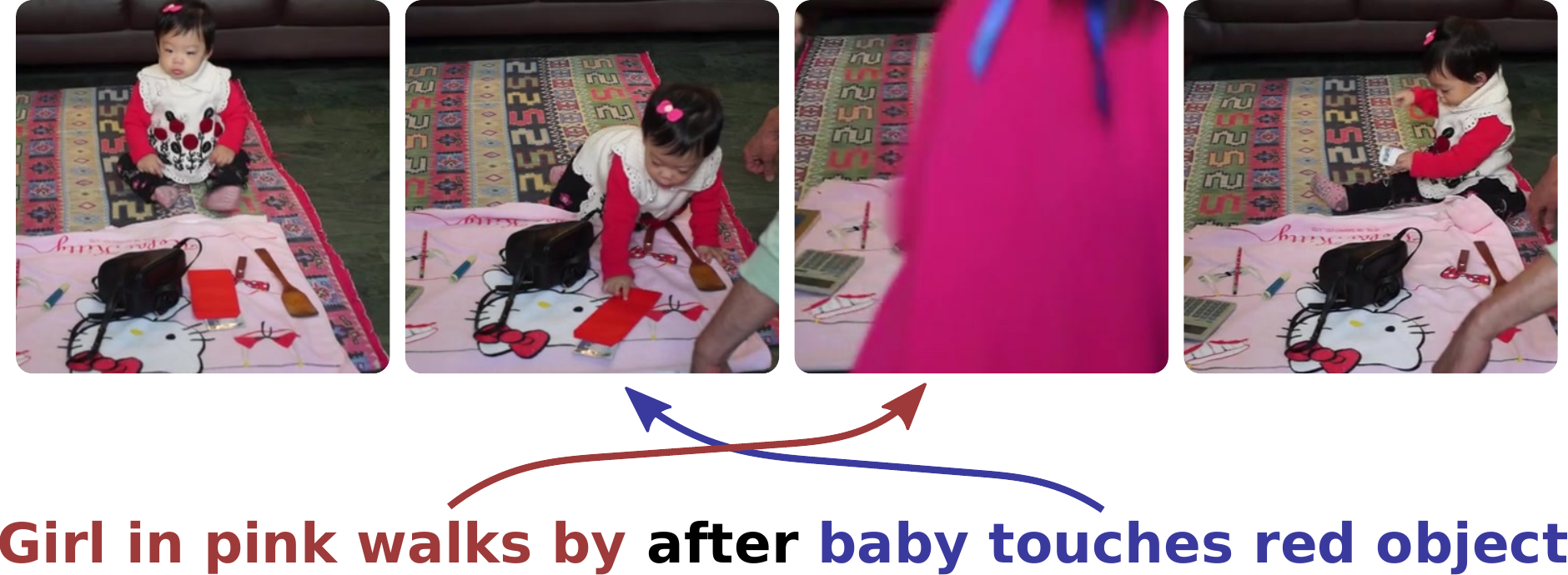}
\vspace{-3mm}
\caption{Example query and video. Queries are often \emph{compositional}, and they impose a \emph{temporal ordering} over their components.}
\label{fig:teaser}
\end{figure}

Like many tasks at the intersection of vision and language, temporal grounding requires that we generalize to both unseen videos \emph{and} unseen queries. Since we make no assumptions about the content of a particular video or query, it is possible for them to depict a completely novel scene or event. This presents a challenge, and one way to overcome it is to leverage our prior knowledge; these modalities have structural properties which can be encoded into a model, which allows us to make better use of training data and generalize more effectively. In our work, we focus on two structural properties inherent to temporal grounding, which we leverage in our model.

The first property is \emph{compositionality}, that is, a single query may be composed of many events. Consider the example in Figure~\ref{fig:teaser}, which contains two events (``walks by'' and ``touches''). Conceptually, there is no limit to the number of events that a query could describe. We can take advantage of this structure by breaking up a query into atomic components, and localizing each of them individually. This makes the problem more manageable; while the full query may be a novel combination of components, it is likely that we will have seen some of these components before.

\begin{figure*}
\centering
\includegraphics[width=0.90\textwidth]{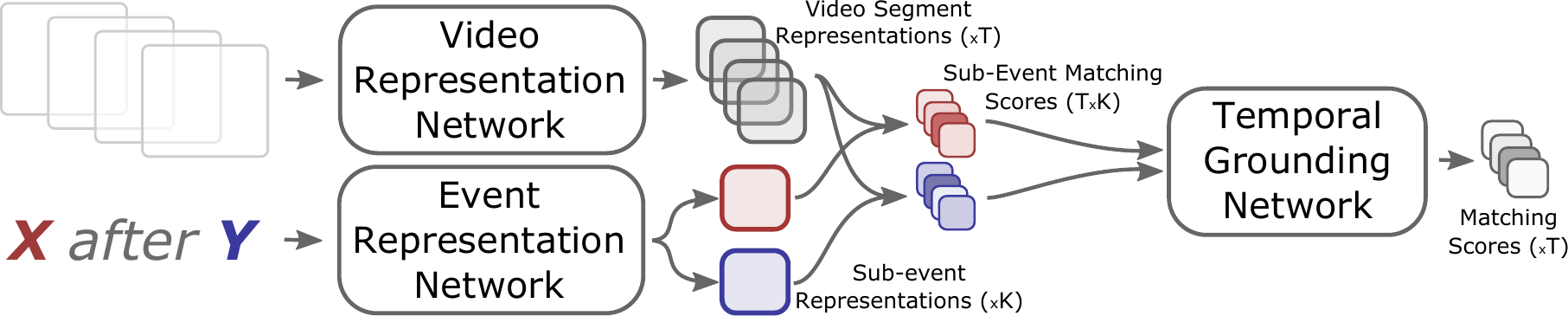}
\vspace{-2mm}
\caption{Overview of our proposed CTG-Net. We represent the input query as several sub-events (depicted as red and blue boxes) using our Event Representation Network (Section~\ref{sec:template}). We match these sub-events to video segments, and then combine and refine these matchings using our Temporal Grounding Network (Section~\ref{sec:grounding}).}
\label{fig:overview}
\end{figure*}

The second property is \emph{temporal ordering}, that is, each query designates a particular ordering of its components. Often, the events must occur in the video in the same order that they appear in the sentence, like in ``\emph{this} then \emph{that}''. But this is not always the case, as in the example in Figure~\ref{fig:teaser}, where the order is reversed. Determining the ordering is non-trivial, as natural language is complex and orderings can be implied from context. However, a model that can determine this ordering can then use it to its advantage, as it allows the model to prune spurious detections which do not match the temporal constraints.

We propose a model, called Compositional Temporal Grounding Network or CTG-Net, which explicitly leverages both compositional and temporal structure for temporal grounding, depicted in Figure~\ref{fig:overview}. Specifically, our model first segments the query into discrete sub-events, leveraging compositionality (Section~\ref{sec:template}). Next, we ground each sub-event in the video, and then refine these groundings to enforce temporal ordering (Section~\ref{sec:grounding}). We apply our model on three temporal grounding datasets: DiDeMo, TEMPO-TL, and TEMPO-HL~\cite{hendricks2017localizing, hendricks2018localizing}, and demonstrate that our approach outperforms state-of-the-art methods (Section~\ref{sec:experiments}). Concretely, we improve Recall@1 from 26.8\% to 28.7\% on TEMPO-TL and from 20.8\% to 21.5\% on TEMPO-HL.

\section{Related Work}

Temporal grounding is a recently-introduced task, and it is often referred to in prior work as moment retrieval or action localization with natural language queries \cite{hendricks2017localizing, gao2017tall}. Prior work on this task generally takes one of two approaches: sliding window or single-shot.

\noindent \textbf{Sliding Window.} In this approach, we embed the query and video segments into the same low-dimensional space, such that the query embedding is similar to the video embedding for the correct segment, and dissimilar for other segments. To ground a query, we slide its embedding across the video and find the most similar segment. Three notable sliding window approaches include Moment Context Networks (MCN) and Moments Localized with Latent Context (MLLC) by Hendricks \etal \cite{hendricks2017localizing, hendricks2018localizing}, Cross-modal Temporal Regression Localizers (CTRL), introduced by Gao \etal \cite{gao2017tall}, and Activity Concepts-based Localizer (ACL), by Ge \etal \cite{ge2019mac}. While these approaches have been successful, they fail to account for compositional events. MLLC makes a step towards this goal, by representing each query as two sub-events, the ``base'' and the ``context''. However, it is restrictive to assume that every event description has exactly two sub-events. Our approach lifts this restriction by allowing any number of sub-events.

\noindent \textbf{Single-Shot.} In this approach, we compare each token in the query to each segment in the video, and then aggregate these comparisons to classify each segment as a correct or incorrect grounding. This includes Temporal GroundNet (TGN), by Chen \etal \cite{chen2018temporally}, Moment Alignment Networks (MAN), by Zhang \etal \cite{zhang2018man}, and Temporal Modular Networks (TMN), introduced by Liu \etal \cite{liu2018temporal}. These approaches do leverage compositionality, but only at the level of individual words, and they generally do not account for temporal relationships. Temporal Modular Networks make significant improvements in terms of compositionality, in that they use the parse tree of the query to gradually aggregate more refined groundings. However, TMN is limited by the use of a fixed dependency parser, while we demonstrate that our method can work both with a fixed parser or in an end-to-end architecture. 

\noindent \textbf{Compositional Representations.} It is well known that events are compositional, and this observation has inspired many prior works in action recognition \cite{rohrbach2012script, gaidon2013temporal, liang2013learning}, captioning \cite{xu2015jointly}, and temporal grounding \cite{liu2018temporal}. Leveraging compositional structures is central to many tasks in computer vision~\cite{felzenszwalb2009object}. To the best of our knowledge, our work is the first to leverage explicit compositional structure for temporal grounding.

\noindent \textbf{Temporal Relationships.} There is a large body of work that addresses temporal relationships in natural language, enabled by corpora with labeled temporal relations~\cite{pustejovsky2003timebank, muller2004annotating}. Prior approaches have used formal logic~\cite{pratt2005temporal, konur2008interval} in addition to machine learning~\cite{mani2006machine, lapata2006learning, chambers2007classifying}. Few papers have approached this problem utilizing the most recent developments in deep learning for NLP, however there has been some recent work on determining temporal ordering from clinical notes \cite{raghavan2014cross, jeblee2018listwise}. We provide a mechanism for incorporating temporal relations which does not rely on hand-designed features or formal logic, by creating a ``position embedding'' for each sub-event. We find that this mechanism, while simple, is effective for reasoning about temporal relations.

\noindent \textbf{Weakly-Supervised Temporal Localization.}
Our work is related to the task of weakly-supervised temporal localization, a well-studied problem in which a system must learn to perform temporal action localization when given only video-level labels \cite{paul2018w,shou2018autoloc, nguyen2018weakly,wang2017untrimmednets,huang2016connectionist}. Similarly, in our work, we do not have individual temporal labels for each atomic sub-event, and must learn to localize these without labels. However, we do have temporal labels for each query. In addition, these prior works use action category labels, as opposed to natural language queries, as the description for each labeled event. Only recently has weakly-supervised temporal localization been studied in the context of natural language queries \cite{tan2019wman}.

\begin{figure}
\centering
\includegraphics[width=0.375\textwidth]{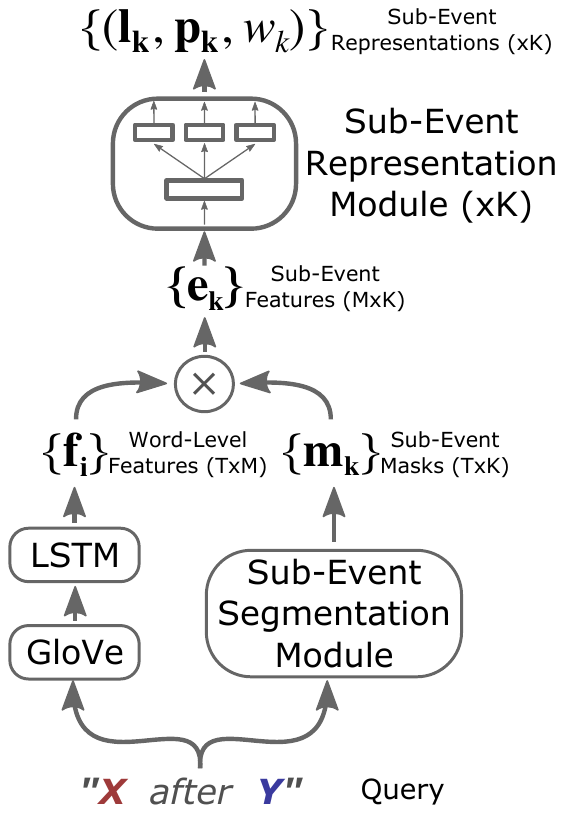}
\vspace{-3mm}
\caption{Event Representation Network with the parser approach. The Sub-Event Segmentation module produces masks that indicate which words belong to each detected sub-event (Section~\ref{sec:masks}). The Sub-Event Representation module produces a triplet representation for each sub-event (Section~\ref{sec:subevent}).}
\label{fig:template}
\end{figure}

\noindent \textbf{Relation to MCN and MLLC.} Our work is a generalization of Moment Context Networks (MCN) \cite{hendricks2017localizing}. Specifically, we use the same encoders for embedding words (Section~\ref{sec:subevent}) and video segments (Section~\ref{sec:video}), as well as the same distance metric to compute matching scores (Section~\ref{sec:grounding}). We generalize MCN by allowing multiple sub-events to each be localized simultaneously, incorporating compositional structure, and later combining and refining their matching scores, incorporating temporal structure. Moments Localized with Latent Context (MLLC)~\cite{hendricks2018localizing} similarly builds off of MCN, but is limited to localizing just two sub-events, and does not include temporal refinement. Our model, therefore, is more flexible and performs well with both simple and complex queries.

\section{Event Representation Network}
\label{sec:template}

We represent each natural language query as a set of one or more sub-events, where the number of sub-events is chosen flexibly depending on the query. We produce this representation using a novel network architecture, which is depicted in Figure~\ref{fig:template} and has two primary components. The first component is the \emph{sub-event segmentation module}, which determines the number of sub-events as well as which words belong to each. The second component is the \emph{sub-event representation module}, which creates a vector representation for each sub-event.

\subsection{Sub-Event Segmentation Module}
\label{sec:masks}

We propose two methods for segmenting a query into sub-events.

\begin{figure}
\centering
\includegraphics[width=0.45\textwidth]{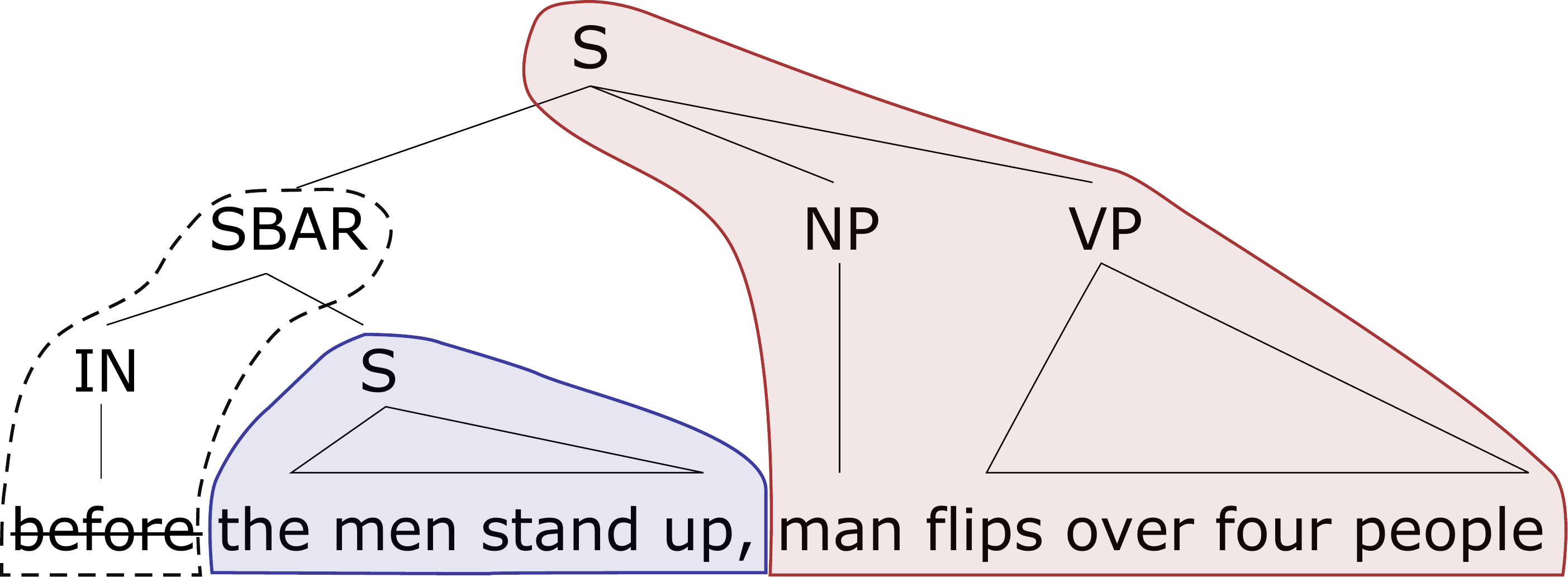}
\vspace{-2mm}
\caption{Result of segmenting sub-events with a parser. This query contains three clauses (with tags S, SBAR, and S). Each word is assigned to the lowest-level clause to which it belongs, and length-1 clauses are discarded. Two sub-events are detected in this example, depicted by the red and blue outlines.}
\label{fig:parser}
\end{figure}

\noindent \textbf{Parser.} In this approach, we use the Stanford Parser \cite{klein2003accurate} to segment the sentence into clauses, and we consider each clause to be a sub-event. We define ``clauses'' to include all clause-level tags in the Penn Treebank (S, SBAR, SINV), as well as fragment tags (FRAG). Since clauses can themselves contain subordinate clauses, we assign each word to the lowest-level clause to which it belongs. For an example of this, see Figure~\ref{fig:parser}. We discard all sub-events that contain only one word, as these are typically due to a conjunction such as ``after,'' which we do not consider to be events.

\noindent \textbf{Bi-directional LSTM.} While the parser approach allows us to segment reasonable sub-events, it imposes some limitations. First, all sub-events are contiguous in the query, and cannot overlap. This introduces an issue when a clause contains a pronoun, as we cannot resolve the reference of the pronoun without the broader context of the sentence. By removing the restrictions of contiguity and overlap, we can replace a pronoun with its referent, making it more easy to localize. Another limitation is that the parser is fixed, and therefore any noise in its output cannot be fine-tuned away during training.

As a flexible and end-to-end learnable alternative to the parser, we propose a simple attention mechanism using a bi-directional LSTM. We feed the word-level features $\mathbf{f_i} \in R^{M} $ (Section \ref{sec:subevent}) into the Bi-LSTM, and then apply $K$ linear classifiers to its output to get a set of attention masks corresponding to $K$ sub-events. Unlike in the parser approach, this always results in an equal number of sub-events. Specifically, we set $K=6$, since this is the maximum number of sub-events identified by the parser in our experiments. However, as we will explain in the following section, each sub-event is later associated with a weight, and it is ignored when its weight is zero. This way, we still can flexibly represent the query as any number of sub-events.

\noindent \textbf{Output.} Mathematically, both segmentation approaches result in a set of masks ${\bf m_k} \in [0, 1]^N$ for $k \in \{1, \dots, K\}$, where $N$ is the number of tokens in the input query and $K \leq 6$ is the number of detected sub-events.

\subsection{Sub-Event Representation Module}
\label{sec:subevent}

To create the sub-event representations, we first compute GloVe embeddings \cite{pennington2014glove} for each word in the query, and then pass these into an LSTM~\cite{hochreiter1997long}. The output of the LSTM is a sequence of feature vectors $\mathbf{f_i} \in \mathbb{R}^M$ for $i \in 1, \dots, N$ where $N$ is the number of words and $M$ is a hyperparameter corresponding to the hidden feature dimension. To convert these word-level features to sub-event-level features, we pool them via a weighted average using the sub-event masks from Section~\ref{sec:masks}. That is, $\mathbf{e_k} = \sum_{i=1}^N m_{ki}\mathbf{f_i}$, resulting in $\mathbf{e_k} \in \mathbb{R}^M$, the feature vector for the $k^\mathrm{th}$ sub-event.

We then create a triplet representation for each sub-event. Specifically, we represent the sub-event $k$ as $(\mathbf{l_k}, \mathbf{p_k}, w_k)$, where $\mathbf{l_k} \in \mathbb{R}^{M_\mathrm{embed}}$ is the language embedding, $\mathbf{p_k} \in \mathbb{R}^{M_\mathrm{pos}}$ is the position embedding, and $w_k \in [0, 1]$ is a scalar weight. The language embedding $\mathbf{l_k}$ is used to co-embed the sub-event features in the same space as the visual features in Section~\ref{sec:video}. The position embedding, $\mathbf{p_k}$, represents the position of the sub-event in time, and is used to enforce temporal consistency in Section~\ref{sec:refinement}. The weight $w_k$ allows the model to ignore sub-events which describe something non-visual, or sub-events which are used purely as context for describing another event (Section~\ref{sec:problem}). Each of these embeddings is created by passing the sub-event features $\mathbf{e_k}$ through a single fully-connected layer. We normalize each $\mathbf{l_k}$ with L2 normalization, and we normalize the weights $w_k$ with a softmax across the K sub-events.

\subsection{Video Representation}
\label{sec:video}

We represent each video segment as an embedding $\mathbf{v_t} \in \mathbb{R}^{M_{\mathrm{embed}}}$, where $\mathbf{t} = (s, e)$ is the start and end-point of the segment. To create the embeddings, we adopt the approach introduced by \cite{hendricks2017localizing}. In this approach, either RGB or Optical Flow frames are passed into a CNN to create frame-level feature vectors. These features are averaged within each segment $\mathbf{t}$ to create \emph{local} features, and averaged across the entire video to get \emph{global} features. We then concatenate the local and global feature vectors along with the start and end-points $(s, e)$ (called temporal end-point features, or TEF) into a $(2D_{\mathrm{video}} + 2)$-length feature vector, where $D_\mathrm{video}$ depends on whether RGB or Optical Flow features are used. This vector is then fed into a 2-layer multi-layer perceptron (MLP) to get the video segment embedding $\mathbf{v_t}$. 

\section{Temporal Grounding}
\label{sec:grounding}

\label{sec:problem}

When queries are compositional, it is not always straightforward to define the expected output of temporal grounding. Consider a query of the form ``\emph{X} after \emph{Y}.'' Should the grounding include the temporal extent of both \emph{X} and \emph{Y}, or only \emph{X}? Some datasets (DiDeMo~\cite{hendricks2017localizing}) encode the answer using the former option, opting to ground the full extent of events described by the query. Other datasets (Tempo-TL, Tempo-HL~\cite{hendricks2018localizing}) however, use the latter, since \emph{Y} in this case is used only to refer to a particular instance of \emph{X}. We propose a method which is agnostic the particular grounding scheme and can be trained end-to-end for either scenario.

Intuitively, our proposed temporal grounding procedure leverages the sub-event representations, as well as their temporal ordering, to find a matching between the query and the video. We first attempt to localize each individual sub-event in the video. Then, we apply a \emph{refinement network} which updates these locations to ensure temporal consistency, and combines them to compute the final grounding. Concretely, this network takes as input the sub-event representations (Section~\ref{sec:subevent}) and video segment representations (Section~\ref{sec:video}), and as output, it produces a score for each time segment $t$, which corresponds to how well that time segment matches the query. 

To locate each sub-event $k$, we compare their embeddings $\mathbf{l_k}$ with the embedding of each video segment $\mathbf{v_t}$. We compute the Euclidean distance between each pair, that is, $d_{kt} = \|\mathbf{l_k} - \mathbf{v_t}\|_2$, where pairs with smaller distances between them are considered better matches. The distance $d_{kt}$ therefore serves as a matching score between sub-event $k$ and video segment $t$.

We then combine these matching scores across sub-events to compute an initial matching score for the entire query. Specifically, we perform a weighted average of the sub-event matching scores, where the weights are given by $w_k$ from the event representation network, that is, $D_t = \sum_{k=1}^{K} d_{kt} w_k$. The weights allow the matching to favor particular sub-events over others, or exclude a sub-event entirely. This is helpful when a sub-event is used only for context, or when a sub-event is not visible in the video.

\subsection{Refinement Network}
\label{sec:refinement}

The initial matching score $D_t$ accounts for compositional structure, but does not account for \emph{temporal} structure. To leverage temporal structure, we propose an additional step, where the matching scores and expected positions of each sub-event are used to update the matching score for the entire query. This phase allows the model to downweight segments that do not match the expected position of each sub-event in the video, which is encoded using the position embedding $\mathbf{p_k}$ (Section~\ref{sec:subevent}).

More precisely, for all video segments $t$, we apply a refinement function $\phi$ which takes as input the matching score $D_t$, the matching score for a particular sub-event $d_{kt}$, the position embedding for that sub-event $\mathbf{p_k}$, and the temporal extent of the segment $t = (s, e)$. We apply the refinement function to each sub-event $k$ and add the results to the matching score $D_t$ to get a refined score $\widetilde{D}_t$. That is,

\begin{equation}
\label{eq:ground2}
\widetilde{D}_t = D_t + \sum_{k=1}^{K} \phi(D_t, d_{kt}, \mathbf{p_k}, t).
\end{equation}

For the refinement function $\phi$, we use a 2-layer MLP. The inputs are concatenated along the feature dimension, which results in an input with dimension $1 + 1 + M_{embed} + 2$. The final matching score $\widetilde{D}$ is used to rank the candidate segments, and as the final grounding we choose the segment that results in the lowest distance $\widetilde{D}$. That is, $\hat{t} = \argmin_t \widetilde{D}_t$.

\subsection{Training}

CTG-Net is fully end-to-end differentiable and can be optimized via gradient descent. We adopt the triplet ranking loss $\mathcal{L}_{\mathrm{triplet}}$ from Hendricks \etal ~\cite{hendricks2017localizing} and apply it directly to the refined matching scores $\mathbf{\widetilde{D}}$. This loss imposes a penalty when the score for an incorrect segment $\widetilde{D}_{t'}$ is lower than that of a correct segment $\widetilde{D}_{t'}$ (recall that lower scores are better), or if they are within some margin $b$, that is, $\mathcal{L}(\widetilde{D}_{t}, \widetilde{D}_{t'}) = \mathrm{max}(0, \widetilde{D}_{t} -  \widetilde{D}_{t'} + b)$. For every positive example, we compare it with two negative examples: one is a segment chosen randomly from the same video $\widetilde{D}_{t_\mathrm{intra}}$, and the other is a segment chosen from a random video at the same timepoint $\widetilde{D}_{t_\mathrm{inter}}$. The loss is a weighted sum of the ranking losses for these two negative examples, that is, $\mathcal{L}_\mathrm{triplet} = \mathcal{L}(\widetilde{D}_{t}, \widetilde{D}_{t_\mathrm{intra}}) + \lambda \mathcal{L}(\widetilde{D}_{t}, \widetilde{D}_{t_\mathrm{inter}})$.

\noindent \textbf{Implementation Details.} We optimize CTG-Net using stochastic gradient descent with a batch size of 120, an initial learning rate of 0.05, and we decay this learning rate by a factor of 10 every 33 epochs (50 for Tempo-HL). We multiply the learning rate by a factor of 10 when updating the LSTM used to create word-level features. We train each network for a maximum of 100 epochs, with early stopping if the validation accuracy plateaus.

We choose hyperparameters from MCN \cite{hendricks2017localizing} where applicable, that is, the word feature dimension $M=1000$, the embedding dimension $M_\mathrm{embed}=100$, the inter-video loss weight $\lambda=0.2$. The remaining hyperparameters, including the dimension of the position embedding $M_\mathrm{pos}=100$ and the size of the hidden layer in the refinement MLP $M_\phi=100$, are selected using the validation set.

As in Hendricks \etal \cite{hendricks2017localizing}, we divide each video into 5-second clips, and consider all segments which consist of one or more contiguous clips. For a 30-second video, this gives 6 possible segments of length 1, 5 possible segments of length 2, and so on. The number of segments for a video with $T$ clips is therefore $T(T+1)/2$.

\noindent \textbf{Late Fusion.} \label{sec:fusion} We use two sets of visual features, one extracted from RGB video frames, and the other extracted from Optical Flow sequences.  For RGB, we pass each frame of the video through a pretrained VGG16 network up to the \emph{fc7} layer \cite{simonyan2014very}, producing a feature vector for each frame. For Optical Flow, we use the penultimate layer of a Temporal Segment Network \cite{wang2016temporal} trained on action recognition.
For each experiment, we train two networks, one for each of the two visual modalities, and then perform a weighted average of their refined correspondences $\lambda_\mathrm{RGB} \widetilde{D}_\mathrm{RGB} + (1-\lambda_\mathrm{RGB}) \widetilde{D}_\mathrm{Flow}$, $\lambda_\mathrm{RGB}=0.3$ to get a fused result. All results include late fusion.

\noindent \textbf{Code.} Our implementation is built in PyTorch~\cite{paszke2017automatic}, and uses GloVE embeddings and visual features (both RGB and Flow) provided by the original creators of the datasets~\cite{hendricks2017localizing, hendricks2018localizing}. Our implementation will be made publicly available.

\section{Experiments}
\label{sec:experiments}

\noindent \textbf{DiDeMo}. Distinct Describable Moments (DiDeMo) is a recent dataset introduced by Hendricks \etal ~\cite{hendricks2017localizing} as a benchmark for temporal grounding. This dataset consists of over 10K unedited videos from Flickr and 40K unique queries. Each query describes a distinct moment in the corresponding video, and its temporal location is annotated independently by four annotators. The queries describe a rich range of events that are not limited to human activities, such as camera and object motion. The dataset includes a high percentage of queries (18.4\%) that include words about temporal relationships, such as ``first'', ``begin'', ``after'', and ``final''. These properties make DiDeMo a realistic benchmark dataset for temporal grounding.

\noindent \textbf{Tempo-TL and Tempo-HL}. Temporal Template Language (Tempo-TL) and Temporal Human Language (Tempo-HL) are two recent temporal grounding datasets which build off of DiDeMo \cite{hendricks2018localizing}. These datasets contain complex queries which are constructed specifically to test compositional grounding, making them an ideal testbed for our method.

Tempo-TL queries are procedurally constructed from pairs of DiDeMo queries. They are constructed using one of five templates: \emph{X before Y}, \emph{Y, before X}, \emph{X after Y}, \emph{Y, after X}, and \emph{X then Y}, depending on the temporal relationship of the two events. While this procedure sometimes results in unnatural sentences, it enables fine-grained evaluation of grounding under specific temporal relationships. Depending on the template, the system must localize the base moment \emph{X} (``before'' and ``after''), or the concatenation of both events \emph{X} and \emph{Y} (``then'').

Tempo-HL queries are constructed by asking each annotator to describe an event \emph{relative} to an existing query from DiDeMo. The resulting queries include all of the same temporal relationships as Tempo-TL, as well as ``while'' relationships, which are not covered by Tempo-TL. Because these queries are rewritten in natural language from scratch, they include coreference statements and a wider range of temporal prepositions. Tempo-HL is therefore a much more realistic and challenging dataset.

\noindent\textbf{DiDeMo+Tempo.} When training on Tempo-TL or Tempo-HL, we also include training examples from DiDeMo, as suggested by the original creators. This is referred to as "DiDeMo+Tempo-TL" or "DiDeMo+Tempo-HL."

\begin{table*}[t]
\begin{center}
\setlength{\tabcolsep}{4pt} % (6pt standard)
\renewcommand{\arraystretch}{1} % (1 standard)
\begin{tabular}{l|c|c|c|c||cc|||c|c|c|c|c||cc|}
& \multicolumn{4}{c||}{Tempo-TL Splits - R@1} & \multicolumn{2}{c|||}{\bf Average} & \multicolumn{5}{c||}{Tempo-HL Splits - R@1} & \multicolumn{2}{c|}{\bf Average} \\
Method & DiDeMo & Before & After & Then & R@1 & R@5 & DiDeMo & Before & After & Then & While & R@1 & R@5 \\ \hline
Prior & 10.7 & 17.9 & 22.4 & 0.0 & 12.7 & 52.6 & 19.4 & 29.3 & 0.0 & 0.0 & 4.7 & 10.7 & 37.6 \\ 
MCN \cite{hendricks2017localizing} & 24.9 & 32.3 & 26.1 & 25.1 & 27.1 & 73.4 & 26.1 & 26.8 & 14.9 & 18.6 & 10.7 & 19.4 & 70.9 \\
TALL \cite{gao2017tall} & 21.0 & 27.1 & 26.3 & 4.8 & 19.8 & 64.7 & 21.8 & 25.9 & 14.4 & 2.5 & 8.1 & 14.6 & 60.7 \\
MLLC \cite{hendricks2018localizing} & 25.9 & 32.0 & 24.3 & 25.0 & 26.8 & 74.0 & 27.4 & {\bf 32.3} & 14.4 & 19.6 & 10.4 & 20.8 & 71.7 \\ \hline
CTG-Net-P &  {\bf 26.8} & 34.1 & {\bf 27.6} & {\bf 26.4} & {\bf 28.7} & {\bf 76.0} & {\bf 27.6} & 27.9 & 16.9 & 18.7 & 10.5 & 20.3 & 71.3 \\ 
CTG-Net-A  & 26.6 & {\bf 35.1} & 26.1 & 23.6 & 27.9 & 75.5 & {\bf 27.6} & 28.6 & {\bf 18.8} & {\bf 20.8} & {\bf 11.6} & {\bf 21.5} & {\bf 72.7} \\
\end{tabular}
\end{center}
\vspace{-3mm}
\caption{\label{tab:tempo} Results on Tempo-TL and Tempo-HL. We compare against prior work, using two variants of our model: with the fixed parser (\emph{CTG-Net-P}), and with the Bi-LSTM attention mechanism (\emph{CTG-Net-A}). For both Tempo-TL and Tempo-HL, we train on the DiDeMo+Tempo-TL and DiDeMo+Tempo-HL training sets, respectively, and evaluate on the test sets. \textbf{Average} refers to the average of each metric across the splits (Section~\ref{sec:experiments}).} \end{table*}

\noindent \textbf{Evaluation.} \label{sec:eval} We adopt the suggested evaluation metrics for both DiDeMo and Tempo. In both datasets, four annotators are each given a video and query and are asked to select the temporal segment which corresponds best with the query. In some cases, there is disagreement between the annotators. To account for disagreement, each metric is computed by comparing the prediction to each of the four annotations, and the annotation that most disagrees with the prediction is discarded. Using this method, we compute three metrics: Recall@1 (R@1), Recall@5 (R@5), and Mean Intersection over Union (mIOU). We note that prior works use the names Rank@1 and Rank@5 when computing these metrics. Our Recall@1 and Recall@5 metrics are equivalent to these.

For DiDeMo, the three metrics are computed for all videos in the dataset. For TEMPO, we use a modification suggested by the original creators \cite{hendricks2018localizing}: we split the dataset into subsets based on the temporal words (``before'', ``after'', ``then'', ``while'') which are present in the queries. We compute the metrics within each subset, and then average the results with equal weight to get a final set of metrics.  This allows us to get a fine-grained understanding of how our model performs under different temporal relationships.

\subsection{Comparison with State of the Art}

\noindent \textbf{Tempo-TL}. In Table~\ref{tab:tempo}, we compare CTG-Net with several prior works and baselines on Tempo-TL. All methods outperform the Prior baseline, which always selects the first video segment. Our model outperforms MCN \cite{hendricks2017localizing} and TALL \cite{gao2017tall}, two sliding-window approaches which do not account for the compositional structures present in Tempo. We also outperform MLLC \cite{hendricks2018localizing}, the previous state-of-the-art on Tempo-TL,  which achieves an average R@1 of 26.8\% compared to 27.9\% for our method with the Bi-LSTM attention mechanism (CTG-Net-A) and 28.7\% for ours with the parser (CTG-Net-P). In terms of average R@5, we also find that CTG-Net-P outperforms MLLC (+2.0\%, from 74.0\% for MLLC to 76.0\% for ours), and performs comparably in terms of mIOU (-0.3\%, from 42.3\% for MLLC to 42.0\% for ours). We perform particularly well on queries containing ``before`` (+3.1\%, from 32.0\% of MLLC to 35.1\% of ours) and and ``after'' (+3.3\%, from 24.3\% of MLLC to 27.6\% of ours), demonstrating our robustness to different temporal relationships. 

\noindent \textbf{Tempo-HL}. In Table~\ref{tab:tempo}, we also perform the same set of comparisons on Tempo-HL. We find that Tempo-HL is more challenging for all temporal words, likely because of the wider range of temporal prepositions and coreferences present in natural language. Our model with the Bi-LSTM (CTG-Net-A) outperforms all prior work on this challenging dataset. We note that our model performs well on the less common temporal relations ``after'' and ``while''. These relations are particularly challenging because they require the model to perform temporal reasoning about sub-events that occur out of order (``after'') and simultaneously (``while''). Our temporal refinement procedure allows the model to take this into account.

We find that, while our model with the fixed parser (CTG-Net-P) outperforms that with the Bi-LSTM attention mechanism (CTG-Net-A) on Tempo-TL, the opposite is true on Tempo-HL. This is likely due to the rigid, procedurally constructed queries in Tempo-TL, which lend themselves well to parsing. Tempo-HL queries, on the other hand, contain more variation in sentence structure, which may not be accounted for by the parser. However, we find that both methods are competitive, demonstrating that compositional structure is useful, regardless of the method of decomposition. Concretely, CTG-Net-A achieves 21.5\% Average R@1 on Tempo-HL, an improvement over MLLC (20.8\%), the previous state-of-the-art. As with Tempo-TL, we also find that CTG-Net outperforms MLLC in terms of R@5 (+1.0\%, from 71.7\% for MLLC to 72.7\% for ours), but performs worse in terms of mIOU (-1.4\%, from 44.6\% for MLLC to 43.2\% for ours). This drop in mIOU is because our model, when compared to MLLC, tends to be over-confident when predicting shorter segments. These shorter segments may contain important components of the event, but this is not reflected in the mIOU metric.

\noindent \textbf{DiDeMo}. In Table~\ref{tab:tempo}, we demonstrate that CTG-Net outperforms MLLC on DiDeMo when trained on both DiDeMo and Tempo-TL (R@1 +1.0\%, from 25.8\% to 26.8\%) and on DiDemo and Tempo-HL (+0.2\%, from 27.4 to 27.6\%). This demonstrates that CTG-Net performs well on the relatively simple queries present in DiDeMo, in addition to the complex queries present in Tempo.

Prior work on DiDeMo does not use the additional training data from Tempo-TL and Tempo-HL, so for fair comparison we additionally report performance of CTG-Net trained on DiDeMo only. When trained on DiDeMo only, CTG-Net-A (27.8\%) outperforms several recent baselines, including Temporal Modular Networks (22.9\%) \cite{liu2018temporal} and Moment Alignment Networks (27.0\%) \cite{zhang2018man}, both of which are competitive recent baselines which leverage compositional reasoning as part of their approach. We additionally achieve competitive results with MCN (28.1\%), MLLC (28.4\%), and Temporal GroundNet (28.2\%) \cite{hendricks2017localizing,hendricks2017localizing,chen2018temporally}. We observe that, suprisingly, CTG-Net-A performs slightly below MCN and MLLC when trained on DiDeMo only, while the opposite is true when we use additional data from Tempo. A simple explanation for this discrepancy is that our model is more complex, and therefore requires more data than is available in the DiDeMo-only setting. In support of this, we observe that our model performs similarly on DiDeMo when trained using the additional data from Tempo-HL (-0.1\%), while other models experience steep \emph{drops} in performance when using this additional data, such as -2.0\% for MCN, and -1.0\% for MLLC. This demonstrates that our model has higher capacity and is more robust to changes in distributions between datasets, both of which are important qualities in practice.

\begin{figure}
\centering
\includegraphics[width=0.485\textwidth]{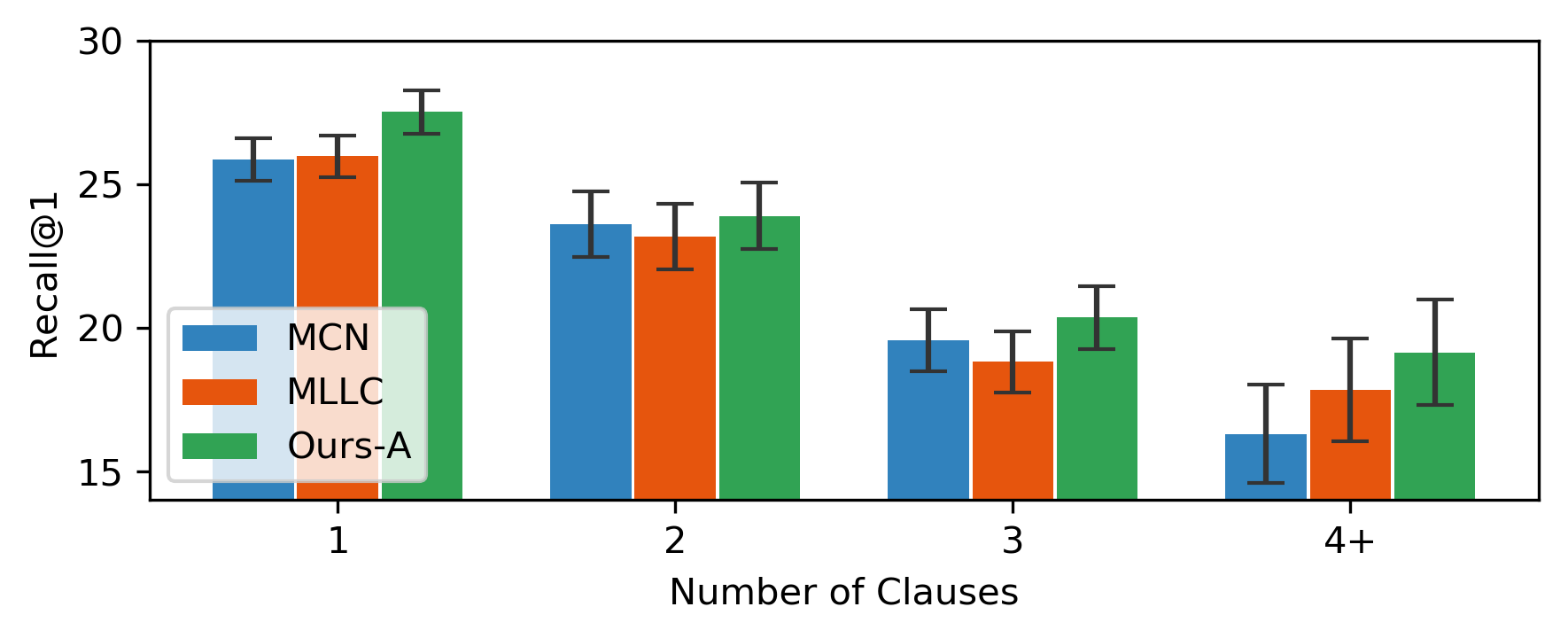}
\includegraphics[width=0.485\textwidth]{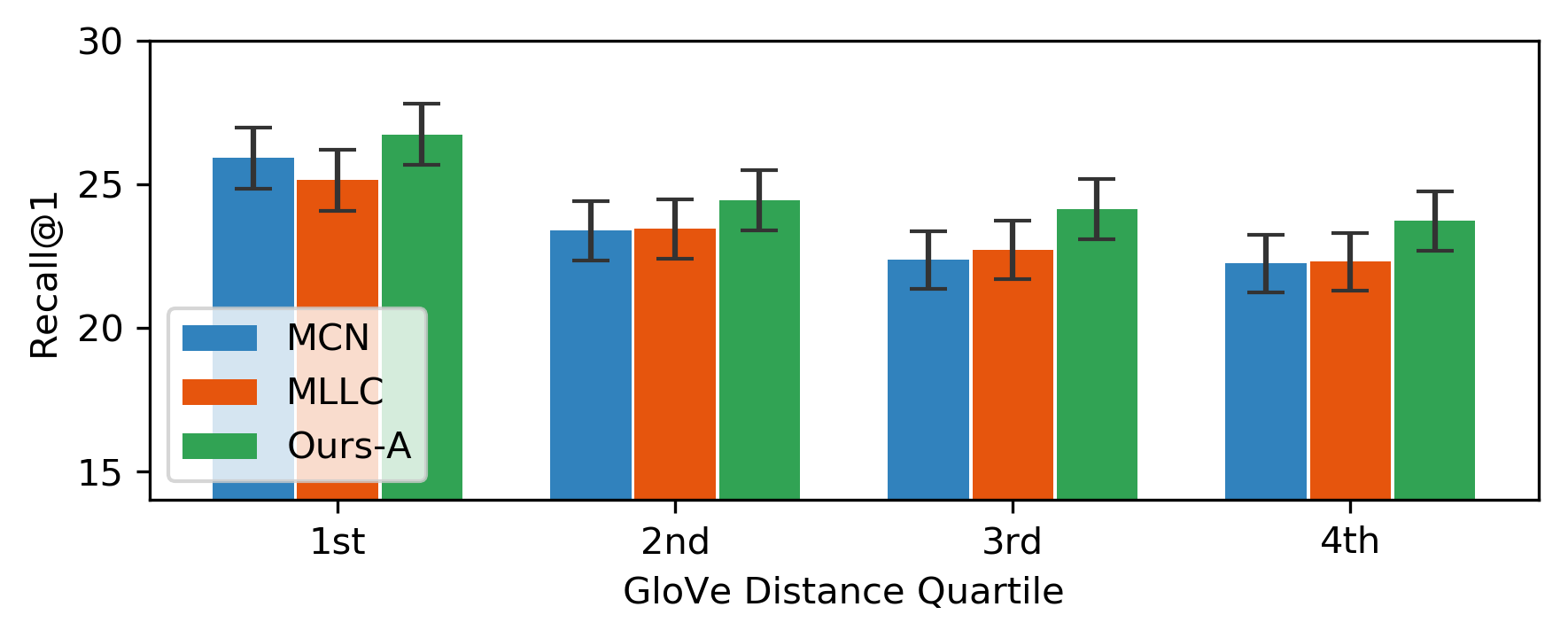}
\vspace{-4mm}
\caption{Recall@1 on DiDeMo+Tempo-HL as a function of query complexity (above) and query novelty (below). Complexity: queries with a high number of clauses are more complex and are therefore more difficult to ground, and our method outperforms prior work at all levels. Novelty: Queries which are dissimilar to previously-seen queries are also more difficult to ground, and our model outperforms prior work at all levels. Error bars depict one standard deviation.}
\label{fig:complexity}
\end{figure}

\begin{table}[t]
\begin{center}
\begin{tabular}{l|ccc}
\hline & \multicolumn{3}{c}{Tempo-HL - \textbf{Average} (Val)} \\
Method & R@1 & R@5 & mIOU \\ \hline
Ours w/o $\mathbf{m_k}$, $\phi$ & 20.75 & 71.91 & 41.68 \\
Ours w/o $\mathbf{m_k}$ & 20.69 & 71.54 & 41.72 \\
Ours w/o $\phi$ & 21.24 & 72.09 & 42.55 \\ \hline
Ours w/o $\mathbf{p_k}$, $w_k$ & 21.49 & 72.46 & 42.31 \\ 
Ours w/o $\mathbf{p_k}$ & 21.37 & 71.68 & 42.58 \\ 
Ours w/o $w_k$ & 21.27 & 71.89 & 42.88 \\ \hline
Ours-A (Full) & {\bf 21.83} & {\bf 72.98} & {\bf 43.25} \\ 
\hline
\end{tabular}
\end{center}
\vspace{-3mm}
\caption{\label{tab:abl} Ablation studies. Top section: to demonstrate the impact of compositional and temporal structure, we remove the sub-event masks $\mathbf{m_k}$ and temporal refinement network $\phi$. Middle section: we remove the position embedding $\mathbf{p_k}$ and weights $w_k$ from the sub-event representations. Our full model outperforms all variants. All numbers are reported on the Tempo-HL validation set.}
\end{table}

\begin{figure*}[t]
\centering
\includegraphics[width=0.75\textwidth]{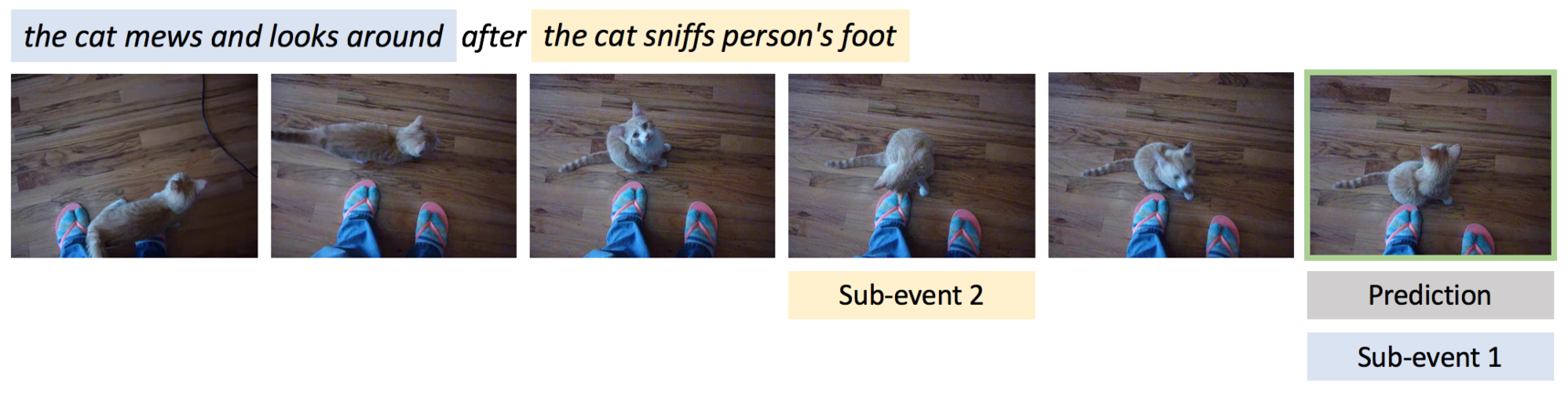}
\includegraphics[width=0.75\textwidth]{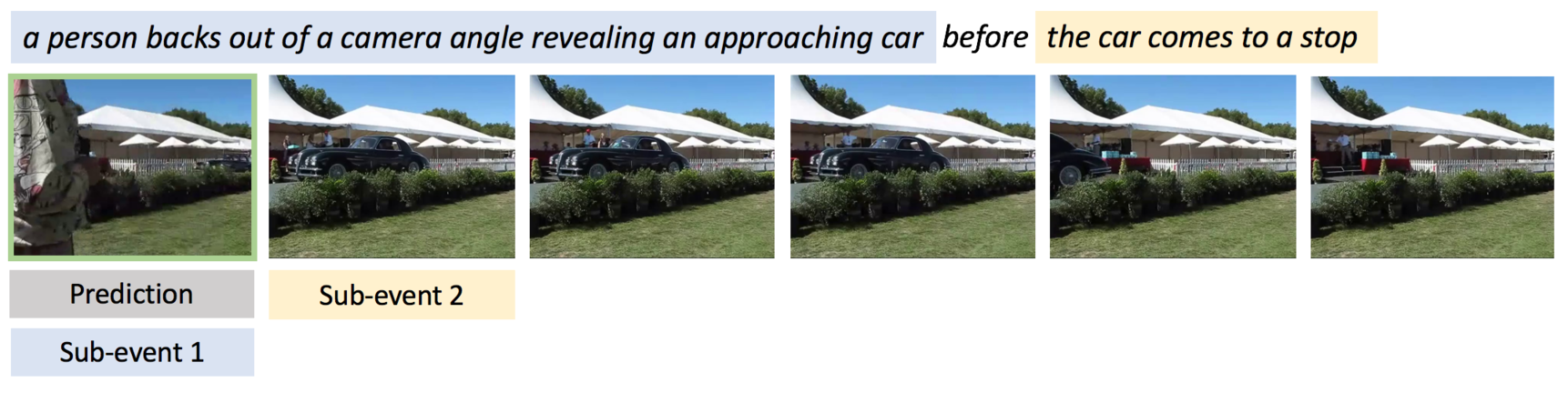}
\includegraphics[width=0.75\textwidth]{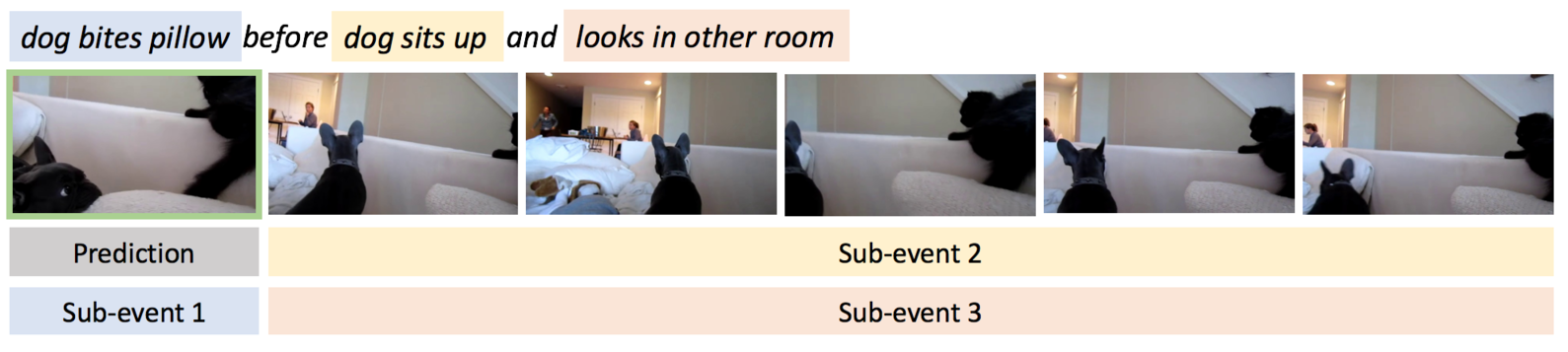}
\vspace{-3mm}
\caption{Examples of sub-event localization on Tempo-TL. The highlighted portions of each query indicate the sub-event masks, and the colored boxes below the video frames indicate the predicted locations of each sub-event (the location with the lowest matching score $d_{kt}$). In both cases, CTG-Net correctly identifies both sub-events, both in the query and in the video.}
\label{fig:qual}
\end{figure*}

\subsection{Complex \& Novel Queries}
\label{sec:complexity}

We expect CTG-Net to generalize well to a broad range of queries. Two challenging areas are \emph{complex} queries, which have many sub-events, and \emph{novel} queries, which are dissimilar to queries seen during training. To test our model in these scenarios, we evaluate its performance on subsets of queries from DiDeMo and Tempo-HL with increasing complexity and novelty.

\noindent \textbf{Complexity}. To measure complexity, we count the number of clauses in each query, using the results of the Stanford Parser~\cite{klein2003accurate}. We divide the queries into categories based on the number of clauses, and we give the performance on each split in Figure~\ref{fig:complexity}. We find that queries with more clauses are more challenging to ground. We also find that our full model (Ours-A) achieves better performance than MCN and MLLC~\cite{hendricks2017localizing, hendricks2018localizing} at all levels, demonstrating that we are able to generalize to both simple and complex queries.

\noindent \textbf{Novelty}. To measure novelty, we compute the average GloVe embedding for each query in the test set, find the most similar average embedding in the training set, and take the Euclidean distance between their embeddings. We consider queries with a high GloVe distance to be more novel. We divide the queries into four categories based on the quartile of the GloVe distance and give the performance in Figure~\ref{fig:complexity}. Novel queries are more challenging, and again we find that our full model achieves better performance at all levels, demonstrating that it is robust to novel queries.

In these experiments, we use Recall@1 over the entire Tempo-HL+DiDeMo test set, \emph{not} the Average R@1 metric (Section~\ref{sec:eval}) that separates by temporal words. This allows us to fairly compare queries based only their complexity and novelty, rather than their source dataset. We use our own implementations of MCN and MLLC, and verify that these implementations achieve similar or better performance than that reported by the original authors: specifically we find that MCN and MLLC achieve Average R@1 scores of 20.6\% and 20.6\% on Tempo-HL, respectively, compared to 19.4\% and 20.8\% in the original work. 

\subsection{Ablation Study}

We perform an in-depth ablation study in Table~\ref{tab:abl}. In these experiments, we demonstrate the contribution of the novel components of our model, namely our use of compositional and temporal structure, and our use of weights and position embeddings in the sub-event representation. In all ablation experiments, we use CTG-Net with attention (Ours-A) trained on DiDeMo+TempoHL, and we report results on the validation set.

\noindent \textbf{Compositional and Temporal Structure}. In the top section of Table~\ref{tab:abl}, we demonstrate the effect of eliminating compositional and temporal reasoning from our model. \emph{Ours w/o $\mathbf{m_k}$} refers to our model without sub-events masks. This model still has the temporal refinement step, but it does not have the benefit of receiving refinement updates from multiple sub-events. \emph{Ours w/o $\phi$} refers to our model without the temporal refinement step, which still uses compositional structure by combining sub-event matching scores. \emph{Ours w/o $\mathbf{m_k}$, $\phi$} has both components removed, and is equivalent to MCN \cite{hendricks2017localizing}. We find that removing both compositional and temporal structure lead to decreased performance (-1.1\%), and that removing compositional structure alone (-1.1\%) is more detrimental than removing temporal structure (-0.6\%). Interestingly, we see no benefit to including temporal refinement without also including sub-event decomposition, demonstrating that temporal refinement leverages updates from multiple sub-events as intended.

\noindent \textbf{Sub-Event Representation}. In the middle section of Table~\ref{tab:abl}, we demonstrate the effect of removing two pieces of our proposed sub-event representation. \emph{Ours w/o} $\mathbf{p_k}$ refers to our model without the position embeddings, which are later used as part of the refinement step. \emph{Ours w/o} $w_k$ refers to our model without sub-event weights, meaning that each sub-event is weighted equally under this scheme.  \emph{Ours w/o} $\mathbf{p_k}$, $w_k$ refers to a model with neither of these components. We find that removing either of these components leads to decreased performance (-0.46\% and -0.56\%), but that removing them both does not lead to further decreased performance, demonstrating that there is still utility in including a sub-event decomposition without these components.

\noindent \textbf{Qualitative Examples.} In Figure~\ref{fig:qual}, we present examples of CTG-Net correctly identifying the locations of individual sub-events in complex queries. For more examples, please refer to the appendix.

\section{Conclusion}
We demonstrate that \emph{compositional} and \emph{temporal} structure are useful for temporal grounding. Specifically, show that event descriptions are \emph{composed} of sub-events, and that they impose an \emph{ordering} on these sub-events. To leverage these structures, we propose Compositional Temporal Grounding Networks (CTG-Net), and show that this model leads to higher performance on challenging datasets when compared with models which do not leverage such structure. We make our code publicly available.

\section{Acknowledgements}
This work is partially supported by King Abdullah University of Science and Technology (KAUST) Office of Sponsored Research (OSR) under Award No. OSR-CRG2017-3405, and by the Toyota Research Institute (TRI).
% TODO: more acknowledgements?

\newpage
\bibliography{main}
\bibliographystyle{ieee_fullname}

\clearpage
\begin{appendices}

\section{Qualitative Examples}

\subsection{Comparison with MCN and MLLC}

\begin{figure*}
\centering
\includegraphics[width=0.9\textwidth]{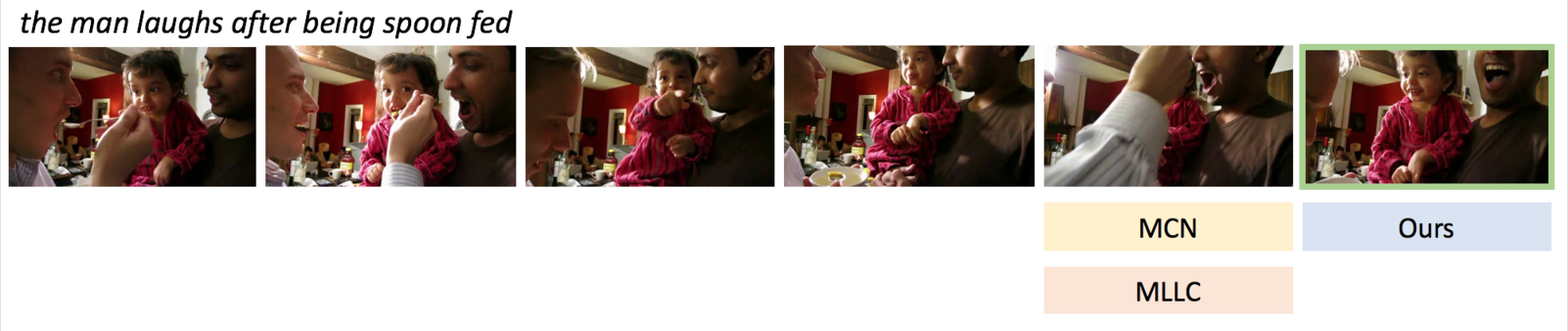}
\includegraphics[width=0.9\textwidth]{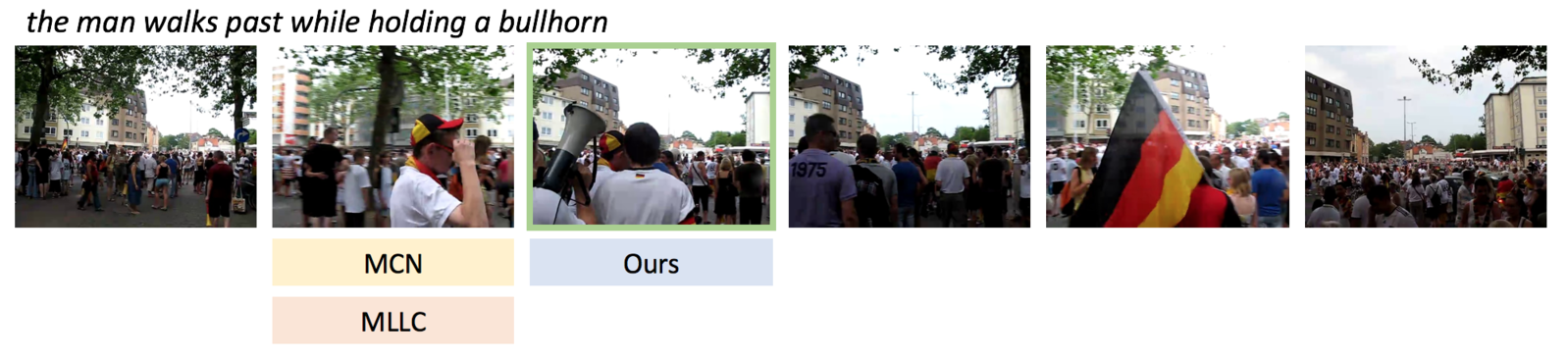}
\includegraphics[width=0.9\textwidth]{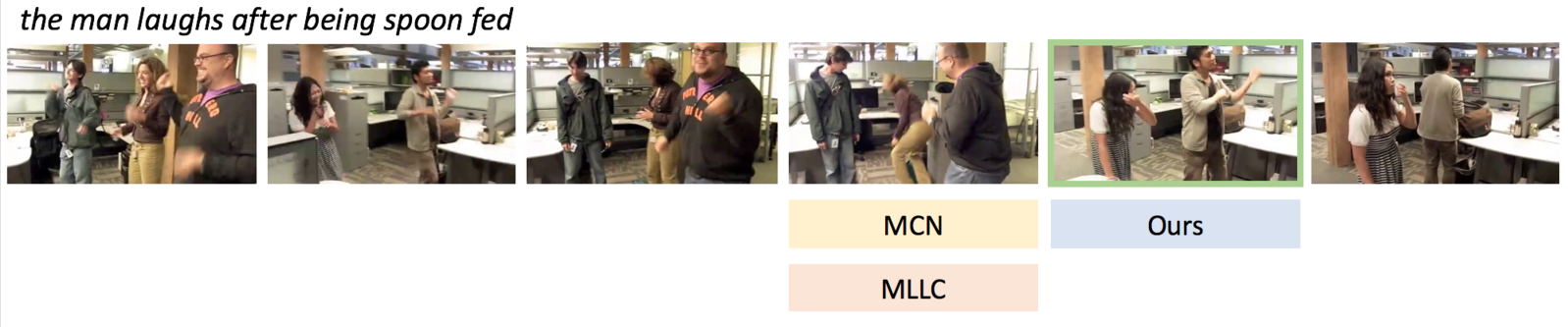}
\includegraphics[width=0.9\textwidth]{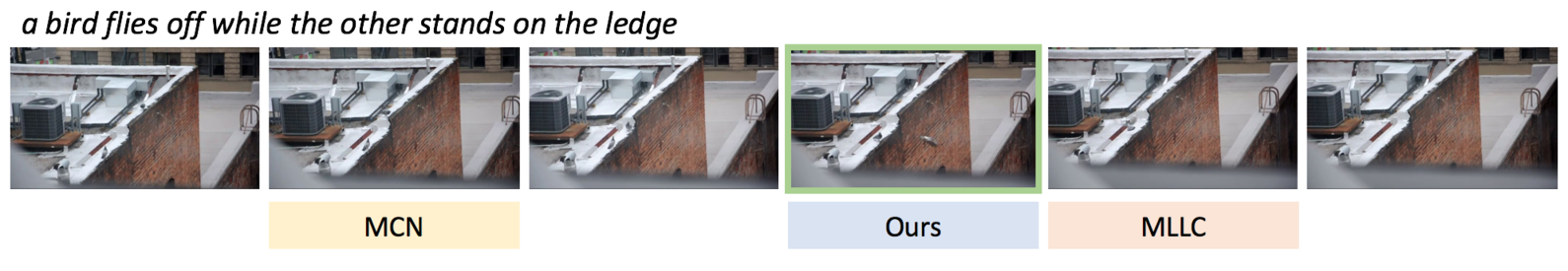}
\includegraphics[width=0.9\textwidth]{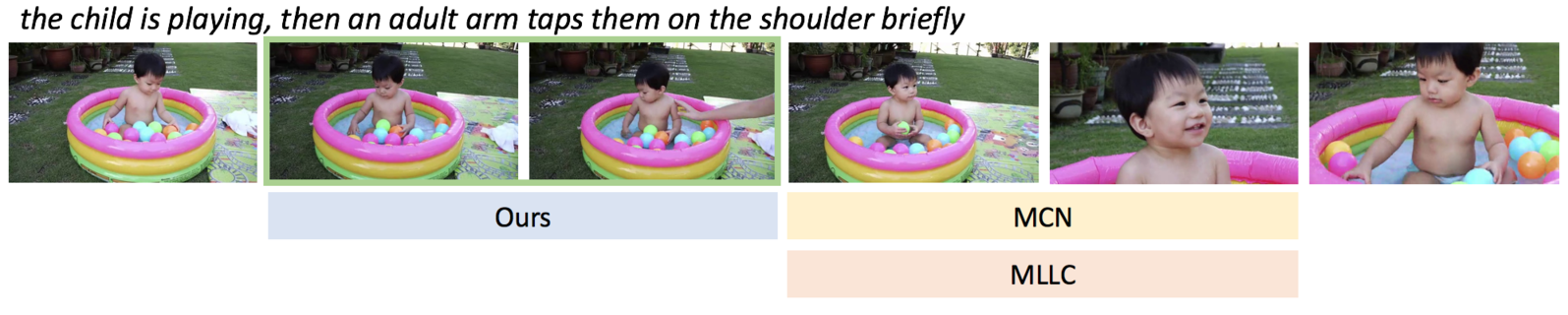}
\caption{Example results from the Tempo-HL and DiDeMo training sets. We compare CTG-Net with the Bi-directional LSTM attention mechanism against results from MCN \cite{hendricks2017localizing} and MLLC \cite{hendricks2018localizing}.}
\label{fig:supp:examples}
\end{figure*}

In Figure~\ref{fig:supp:examples}, we provide examples of challenging instances from Tempo-HL and DiDeMo, and show the temporal segment chosen by our model compared to prior work. Our model reliably localizes difficult queries that are missed by both MCN \cite{hendricks2017localizing} and MLLC \cite{hendricks2018localizing}.

\subsection{Examples of Compositional Grounding}

\begin{figure*}
\centering
\includegraphics[width=0.9\textwidth]{fig/examples/subevent_pos1.png}
\includegraphics[width=0.9\textwidth]{fig/examples/subevent_pos2.png}
\includegraphics[width=0.9\textwidth]{fig/examples/subevent_pos3.png}
\includegraphics[width=0.9\textwidth]{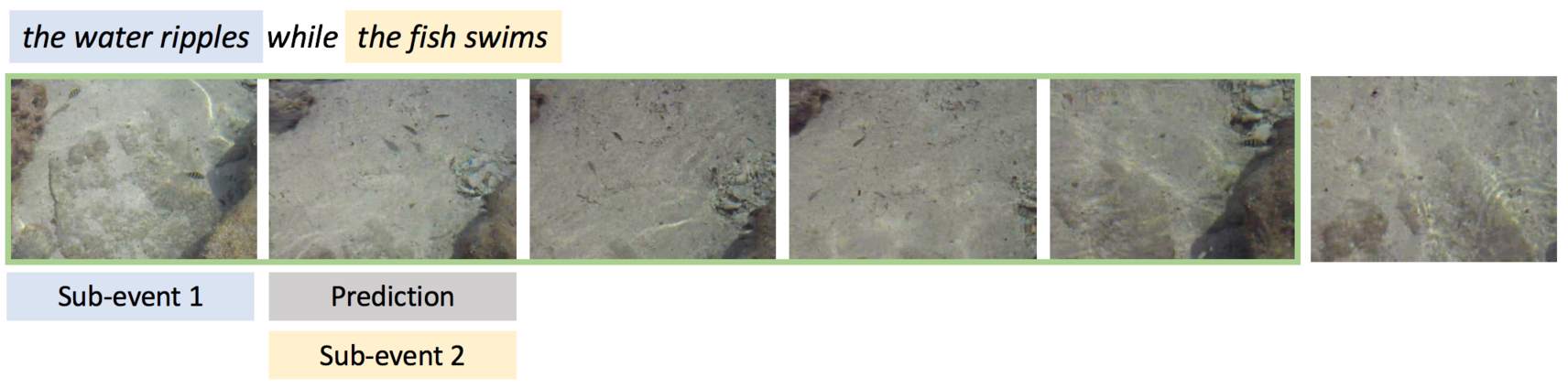}
\includegraphics[width=0.9\textwidth]{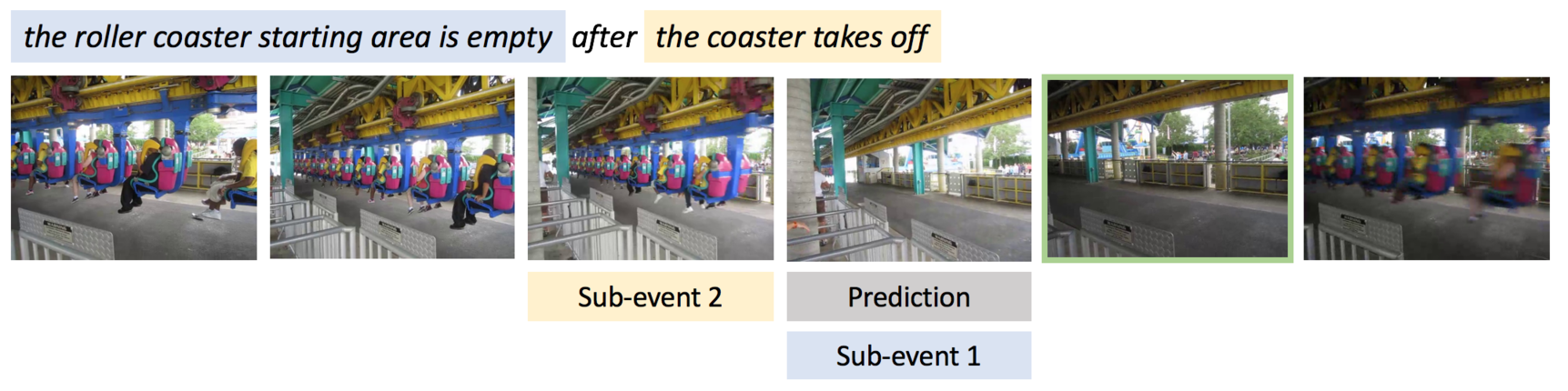}
\caption{Example results of temporally grounded sub-events from Tempo-HL and DiDeMo. We CTG-Net model with the dependency parser, and show the individual temporal groundings before combination and refinement. The top 3 examples depict examples of accurate compositional groundings, and the bottom 2 depict incorrect groundings.}
\label{fig:supp:grounding}
\end{figure*}

In Figure~\ref{fig:supp:grounding}, we provide examples of instances from Tempo-HL and DiDeMo, and show the temporal segments that we identify as sub-events. We show that CTG-Net identifies sensible sub-events, and later uses these to create a final grounding.

\subsection{Examples Before and After Refinement}

\begin{figure*}
\centering
\includegraphics[width=0.9\textwidth]{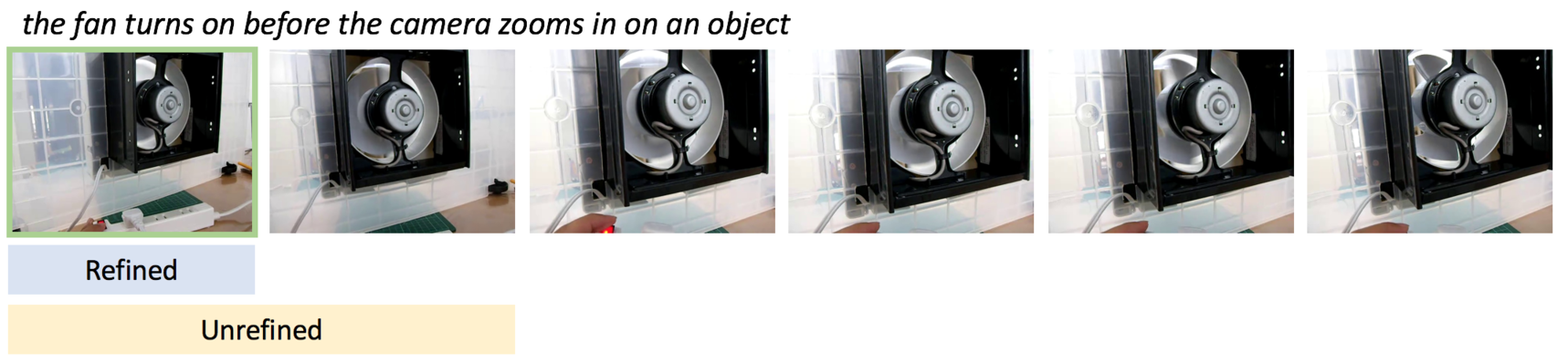}
\includegraphics[width=0.9\textwidth]{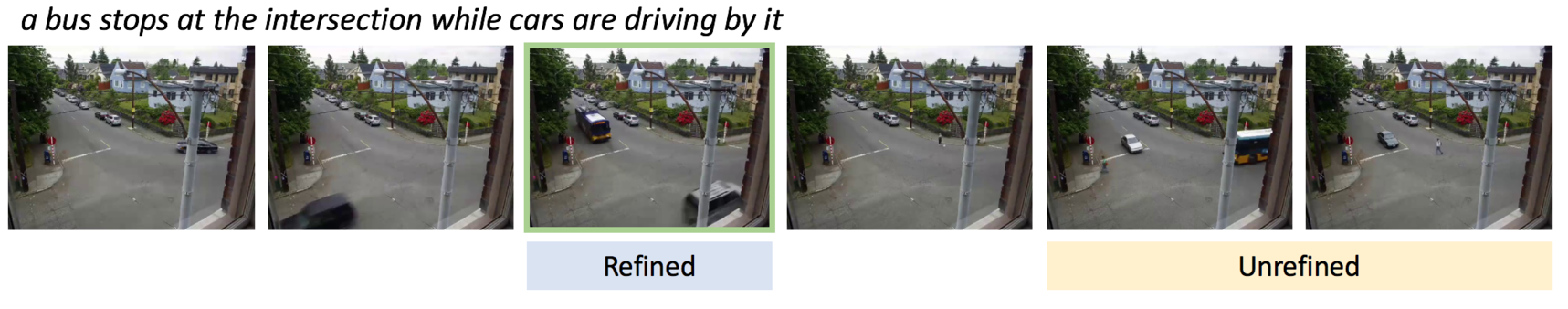}
\includegraphics[width=0.9\textwidth]{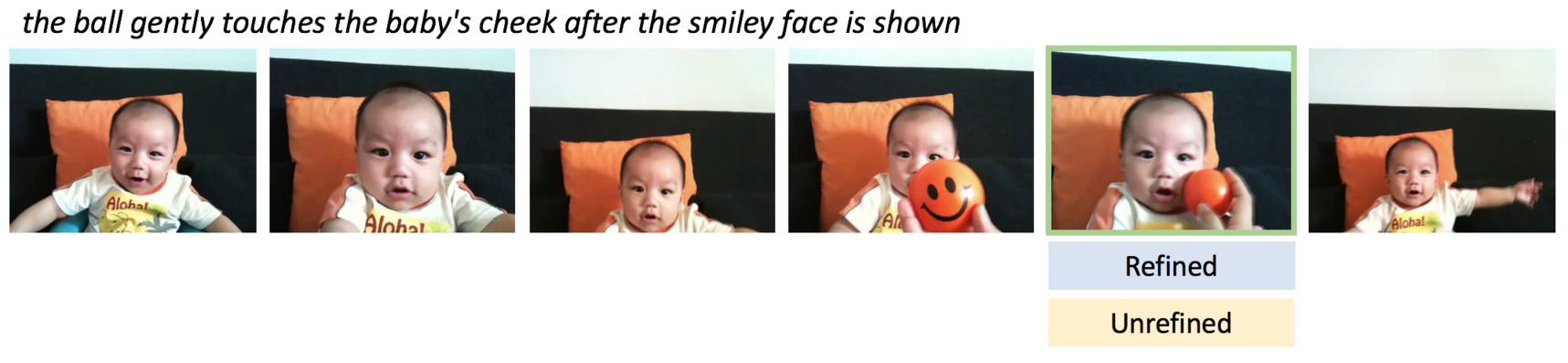}
\includegraphics[width=0.9\textwidth]{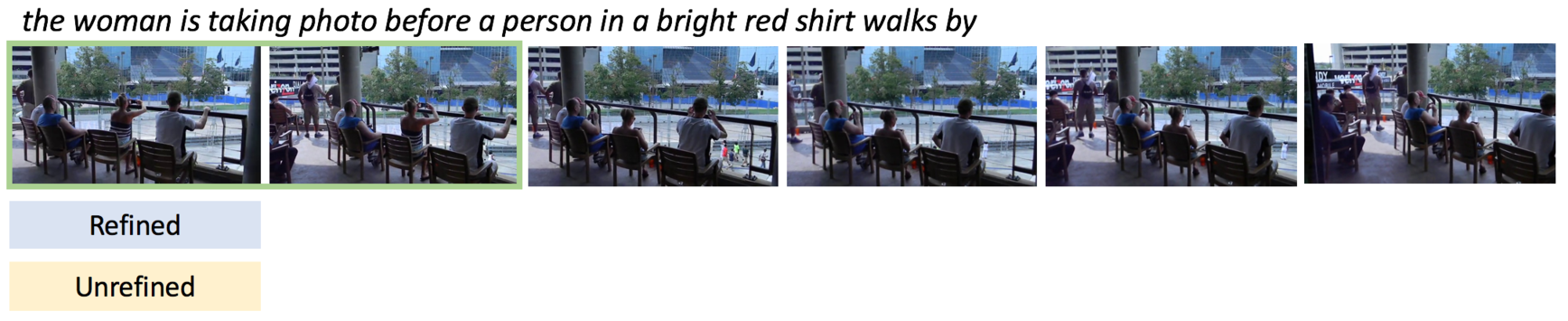}
\caption{Example results of CTG-Net before and after the refinement step, shown on Tempo-HL and DiDeMo. In the top 3 examples, we see that the refinement step is able to improve the prediction. In the bottom example, we show a difficult case where the refinement process leads to incorrect predictions.}
\label{fig:supp:refinement}
\end{figure*}

In Figure~\ref{fig:supp:refinement}, we provide examples of instances from Tempo-HL and DiDeMo, grounded using CTG-Net before and after the refinement step. Refinement improves groundings.

\section{Dependency Parser}

\subsection{Example Segmentations}

\begin{figure*}
\centering
\includegraphics[width=0.65\textwidth]{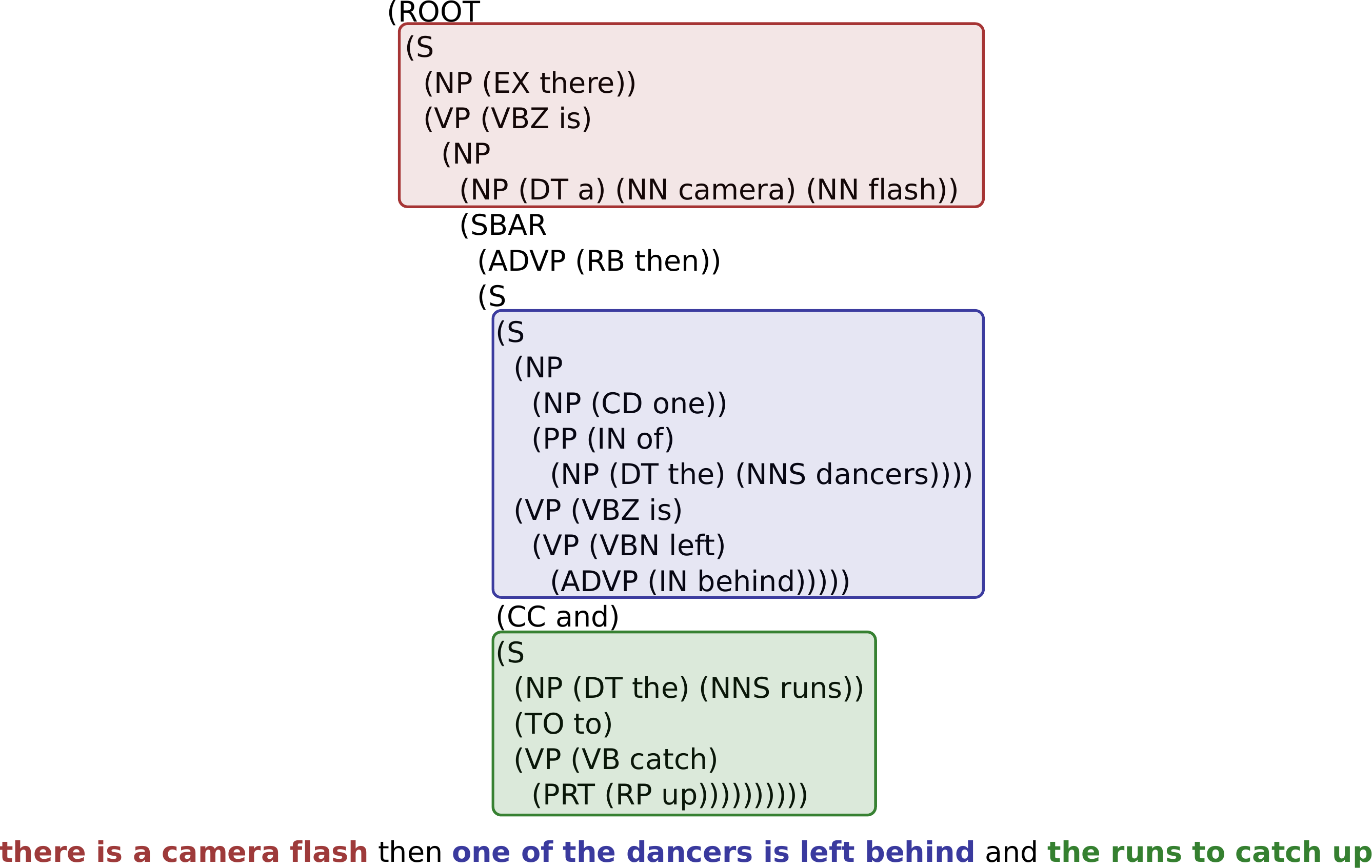}
\vspace{5mm}
\\
\includegraphics[width=0.6\textwidth]{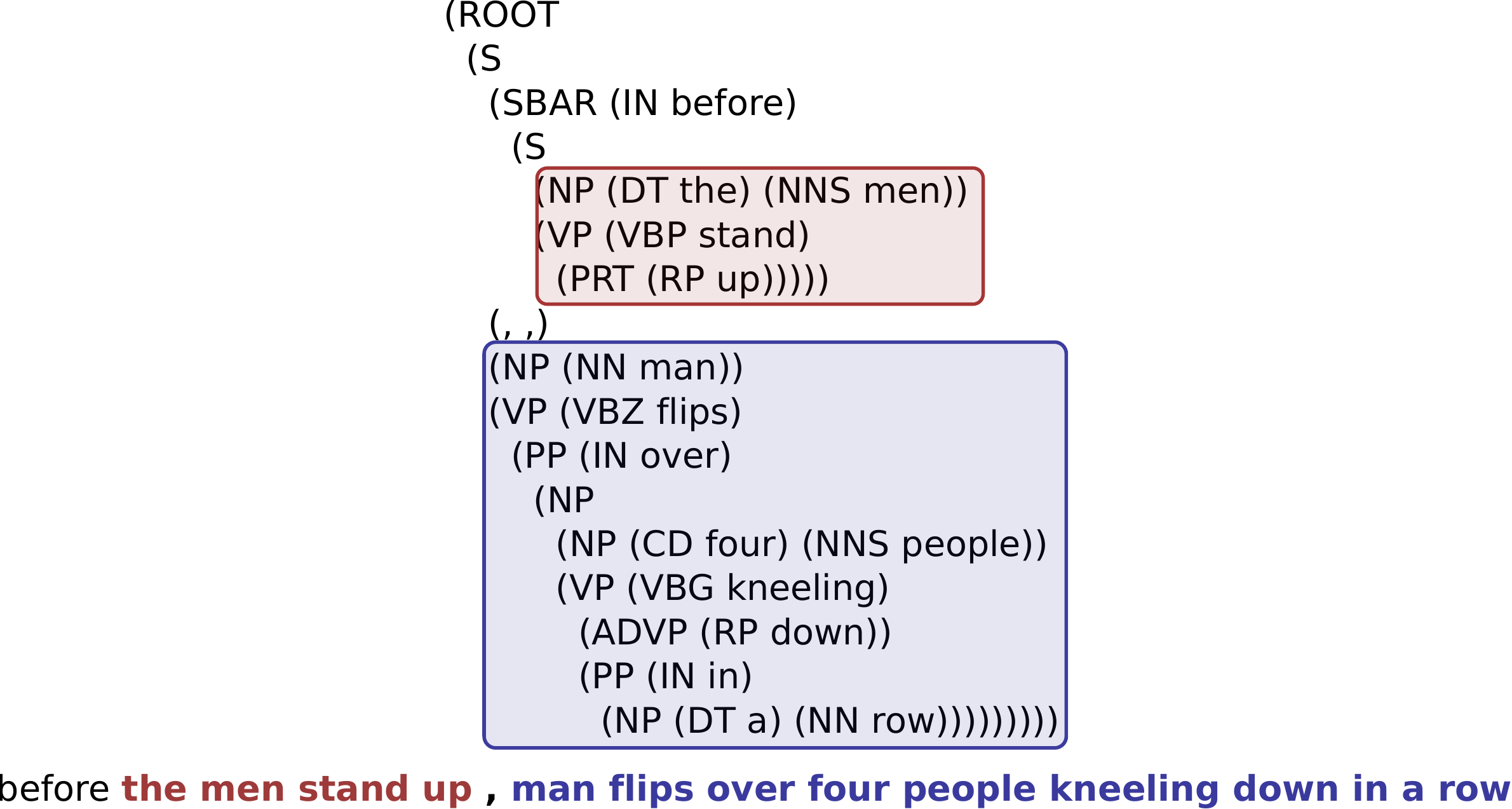}
\vspace{5mm}
\\
\includegraphics[width=0.7\textwidth]{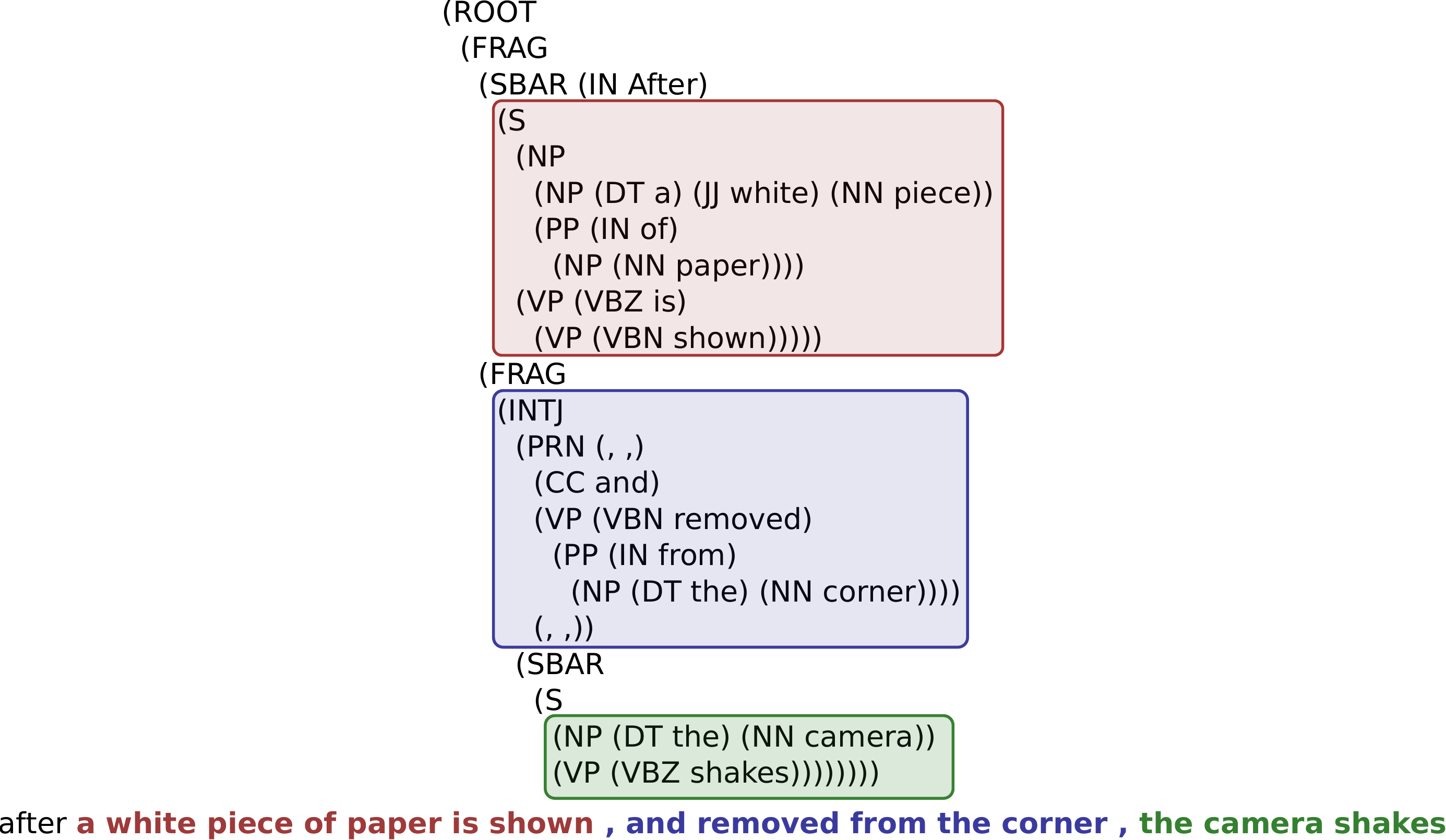}
\caption{Example sub-event segmentations produced by the dependency parser.}
\label{fig:supp:parse}
\end{figure*}

In Figure~\ref{fig:supp:parse}, we show examples of sub-events identified by CTG-Net with the parser approach. This model identifies sensible sub-events, which can each be grounded in the video.

\subsection{Distribution of Sub-Events}

\begin{figure*}
\centering
\includegraphics[width=0.32\textwidth]{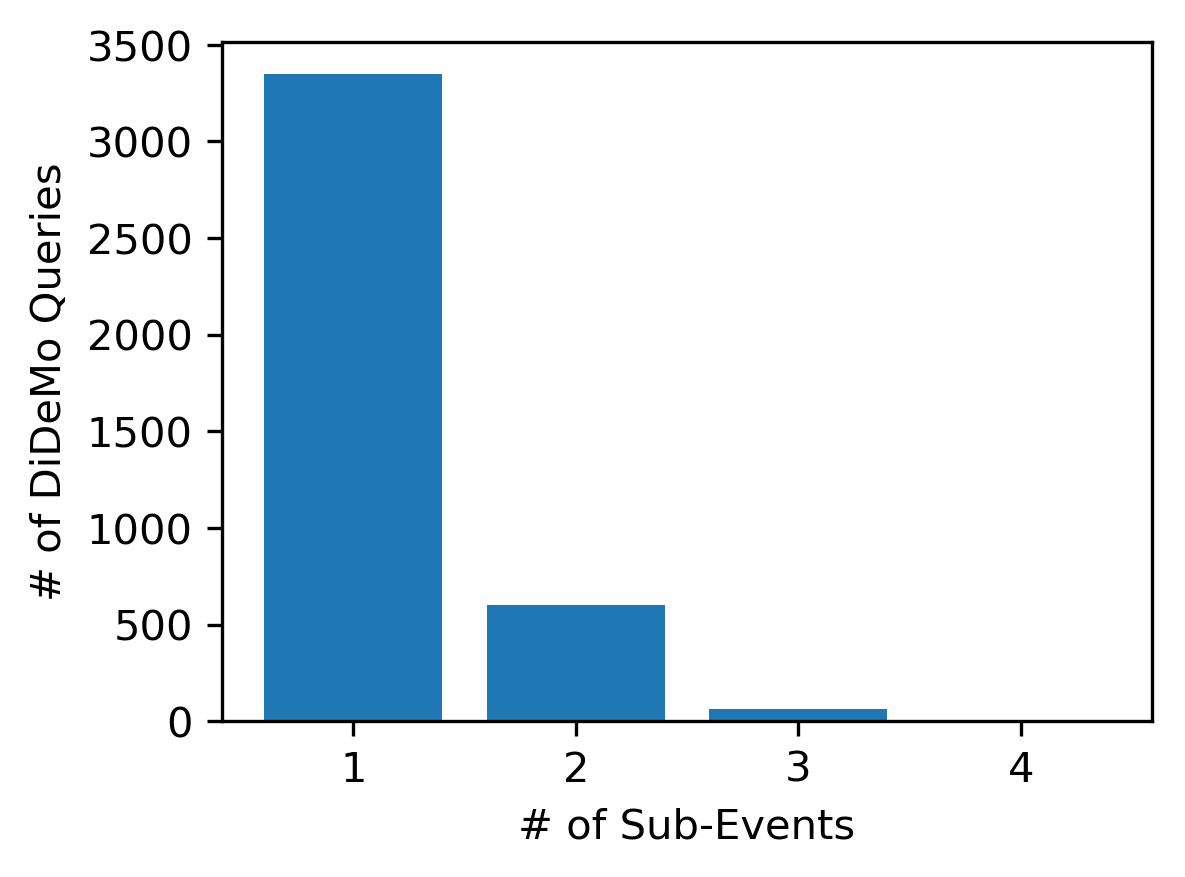}
\includegraphics[width=0.32\textwidth]{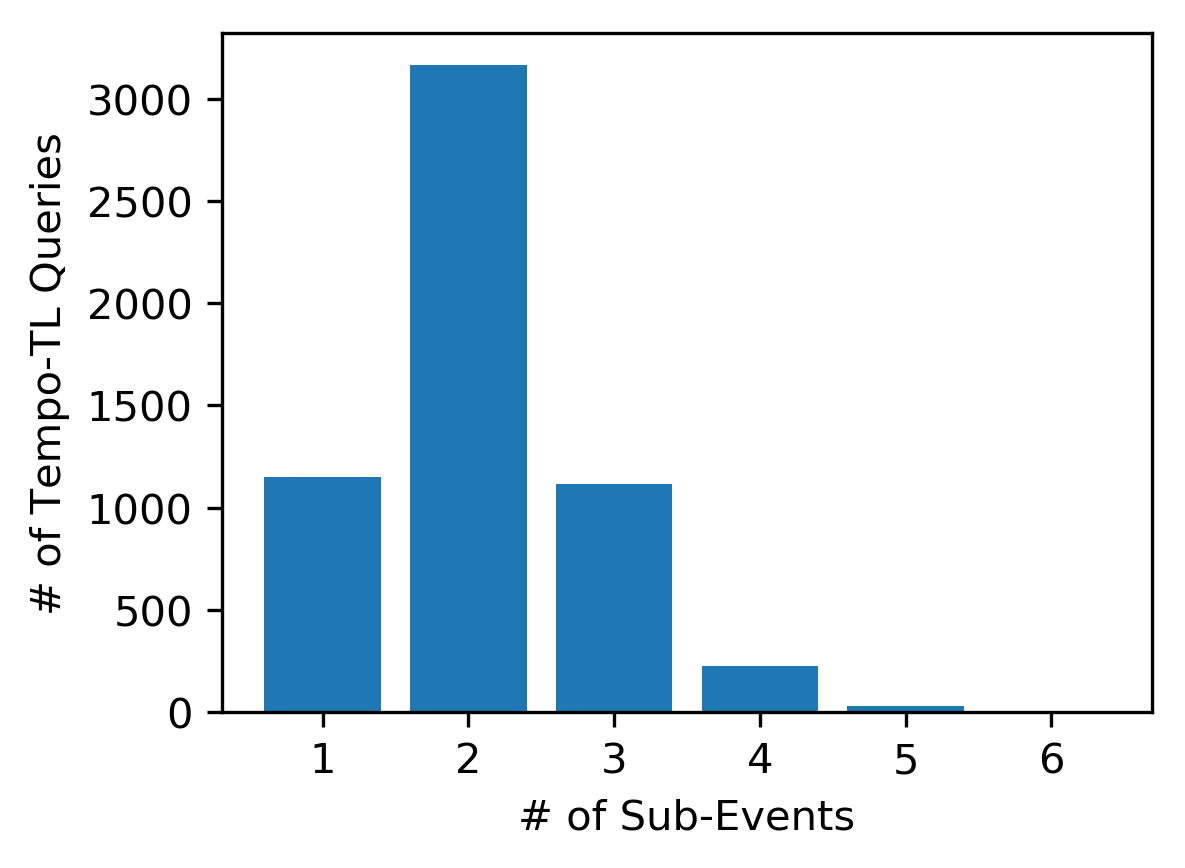}
\includegraphics[width=0.32\textwidth]{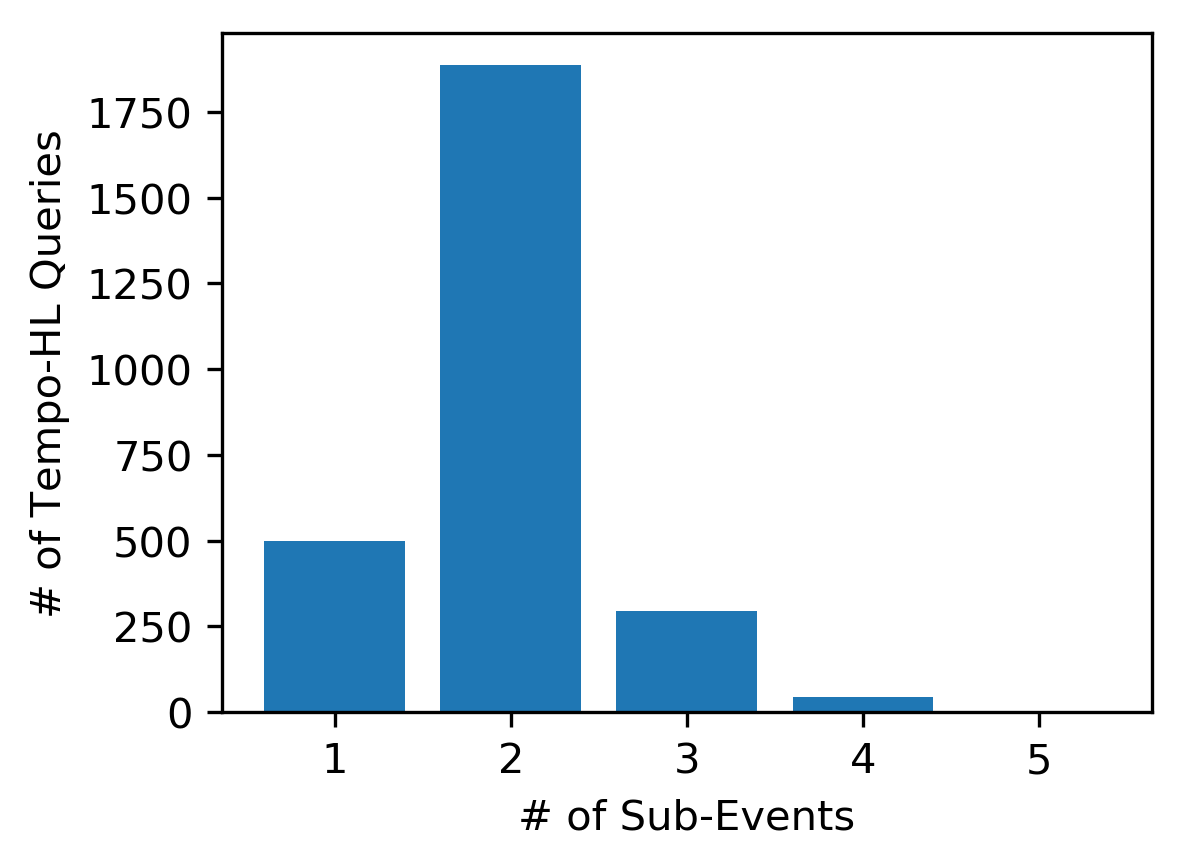}
\caption{Distribution of number of sub-events in DiDeMo (left), Tempo-TL (middle) and Tempo-HL (right), as identified by our dependency parser approach. DiDeMo queries are typically shorter and contain few sub-events. Tempo-TL and Tempo-HL are strongly biased towards two sub-events per query.}
\label{fig:supp:parsehist}
\end{figure*}

In Figure~\ref{fig:supp:parsehist}, we show the distribution of the number of sub-events identified by the parser in each of the three datasets: DiDeMo, Tempo-TL, and Tempo-HL.

\end{appendices}

\end{document}